\documentclass[10pt,twocolumn,letterpaper]{article}

\usepackage{iccv}
\usepackage{times}
\usepackage{epsfig}
\usepackage{graphicx}
\usepackage{amsmath}
\usepackage{amssymb}

\usepackage{amsmath,amsfonts,epsfig,graphicx,subfigure,url,multirow,booktabs,collcell}
\usepackage{array,multirow,graphicx,tablefootnote,threeparttable}
\usepackage[detect-none]{siunitx}
\newcommand{\tabincell}[2]{\begin{tabular}{@{}#1@{}}#2\end{tabular}}

\usepackage[breaklinks=true,bookmarks=false]{hyperref}

\iccvfinalcopy 

\def\iccvPaperID{292} 

\begin{document}

\title{Learning long-term dependencies for action recognition with a biologically-inspired deep network}

\author{Yemin Shi$^{1,2}$, Yonghong Tian$^{1,2}$\thanks{Corresponding author: Yonghong Tian (email: yhtian@pku.edu.cn).}, Yaowei Wang$^{3*}$, Tiejun Huang$^{1,2}$\\
$^{1}$ National Engineering Laboratory for Video Technology,
School of EE\&CS, \\
Peking University, Beijing, China\\
$^{2}$ Cooperative Medianet Innovation Center, China\\
$^{3}$ School of Information and Electronics,\\
Beijing Institute of Technology, Beijing, China
}


\maketitle

\begin{abstract}
Despite a lot of research efforts devoted in recent years, how to efficiently learn long-term dependencies from sequences still remains a pretty challenging task. As one of the key models for sequence learning, recurrent neural network (RNN) and its variants such as long short term memory (LSTM) and gated recurrent unit (GRU) are still not powerful enough in practice. One possible reason is that they have only feedforward connections, which is different from the biological neural system that is typically composed of both feedforward and feedback connections. To address this problem, this paper proposes a biologically-inspired deep network, called shuttleNet\footnote{Our code is available at \url{https://github.com/shiyemin/shuttlenet}}. Technologically, the shuttleNet consists of several processors, each of which is a GRU while associated with multiple groups of cells and states. Unlike traditional RNNs, all processors inside shuttleNet are loop connected to mimic the brain's feedforward and feedback connections, in which they are shared across multiple pathways in the loop connection. Attention mechanism is then employed to select the best information flow pathway. Extensive experiments conducted on two benchmark datasets (i.e UCF101 and HMDB51) show that we can beat state-of-the-art methods by simply embedding shuttleNet into a CNN-RNN framework.
\end{abstract}

\section{Introduction}

Deep neural networks (DNNs) have achieved great success in recent years as they are able to learn complex features and patterns from data. A typical DNN has multiple nonlinear layers which are connected with feedforward connections. In order to gain better performance, very deep structure must be considered to construct a DNN, which will then leads to massive amount of parameters and high risk of over-fitting.

In biological neural system, the visual areas of the brain are interconnected in a complex pattern of feedforward, lateral, and feedback pathways \cite{felleman1991distributed,macknik2009role}. Feedback connections are ubiquitous throughout the cortex, and subcortical regions in ascending hierarchical pathways also receive a large amount of feedback from cortical areas \cite{ericsir1997relative,sherman2002role}. This fact motivates us that DNNs may benefit a lot from imitating the biological neural system by introducing lateral or feedback connections.

Following the similar idea, recurrent neural networks (RNNs) introduce lateral connections to the temporal domain to condition their present state on the entire history of inputs. Because of the temporal lateral connection mechanism, RNNs are able to capture long-term dependencies in sequential data over an extended period of time. Moreover, RNNs have been theoretically proved to be a Turing-complete machine, indicating that they can be used to approximate any functions \cite{siegelmann1992computational}. As one variant of RNNs, long short term memory (LSTM) \cite{hochreiter1997long} is proposed to solve the gradient vanishing and exploding problems. When unfolded in time, LSTMs are equivalent to very deep neural networks that share model parameters and receive the input at each time step. The parameter-sharing mechanism guarantees that there are not too many parameters and consequently the network is trainable. Many works \cite{donahue2015long,shi2017sequential,yue2015beyond} have proved the effectiveness of LSTM on the action recognition task from video sequences.

In this paper, we propose a new kind of biologically-inspired deep neural network, called shuttleNet, which is composed of both feedforward and feedback connections. Technologically, the shuttleNet consists of $N$ processors and $N\times D$ groups of cells and states. In our model, each processor is a simple Gated Recurrent Unit~\cite{cho2014learning}, typically containing a group of weight matrices and associated with $D$ groups of cells and states. All the $N$ processors are organized as a ring, each of which clockwise connects to the next processor with a stride of $K$. If an input $x$ is fed into the $n^{th}$ processor, it will return to the original processor after $\frac{N}{K}$ steps (typically $D \geq \frac{N}{K}$). This process forms a loop connection which can be viewed as a combination of both feedforward and feedback connections. In the network, $x$ will be fed into all the $N$ processors and be passed through $D$ steps, consequently leading to $N$ pathways. Moreover, these processors can be shared in different pathways while having standalone states at each step. After that, the attention mechanism \cite{bahdanau2014neural} is employed to select the best pathway, suppose that each pathway corresponds to one potential prediction. In short, the network works in a shuttle way, thus the name shuttleNet.

We evaluate the proposed shuttleNet for action recognition task on two benchmark datasets, namely UCF101 and HMDB51. Experimental results show that the shuttleNet outperforms LSTMs and GRUs remarkably. By simply embedding our shuttleNet into a CNN-RNN network, we can beat most of the state-of-the-art action recognition methods on both datasets.

The rest of the paper is organized as follows: In section \ref{sec:related_work}, we review the related work. We will briefly introduce the biological background and our motivation in section \ref{sec:background}. The proposed shuttleNet is presented in section \ref{sec:shuttleNet}. Experimental results are discussed in section \ref{sec:experiments}. Finally, section \ref{sec:conclusion} concludes this paper.

\section{Related work}\label{sec:related_work}

Basically, action recognition aims at categorizing the actions or behaviors of one or more persons in a video sequence. Two-stream ConvNets \cite{simonyan2014two} is widely recognized as the first successful deep learning framework for action recognition. It extracts the spatial and temporal characteristics in one framework, and trains the standalone CNNs for two streams separately. Wang \textit{et al.} \cite{wang2015towards} also successfully trained very deep two-stream ConvNets on the UCF101 dataset. Similarly, trajectory-pooled deep-convolutional descriptor (TDD) was proposed by Wang \textit{et al.} \cite{wang2015action}, which shares the merits of both hand-crafted features such as dense trajectories \cite{wang2011action, wang2013action} and deeply-learnt features. However, two-stream ConvNets did not capture and utlize the long-term dependence in the network.

Hierarchical recurrent neural network \cite{el1995hierarchical} is one of the earliest works which attempted to improve the efficiency of capturing long term dependency. Long short term memory (LSTM) \cite{hochreiter1997long}, the most successful approach to deal with vanishing gradients, was proposed by Hochreiter and Schmidhuber. Basically, LSTM relies on a fantastic structure made of gates to control the flow of information to the hidden neurons. Peephole LSTM \cite{gers2000recurrent} adds peepholes to some gates so as to allow them look at the cell state. Gated recurrent unit (GRU), introduced by Cho \textit{et al} \cite{cho2014learning}, is a slightly more dramatic variation on the LSTM, which combines the forget and input gates into a single update gate and merges the cell state and hidden state. Mikolov \textit{et al.} \cite{mikolov2014learning} proposed to add a hidden layer to RNNs and make the weight matrix close to identity.

It should be noted that LSTMs were introduced to model long-term actions for action recognition recently. Yue-Hei \textit{et al.} \cite{yue2015beyond} and Donahue \textit{et al.} \cite{donahue2015long} proposed their own recurrent networks respectively by connecting LSTMs to CNNs. Wu \textit{et al.} \cite{wu2015modeling} achieved the state-of-the-art performance by connecting CNNs and LSTMs under the hybrid deep learning framework. Shi \textit{et al.} \cite{shi2015learning,shi2017sequential} also introduced their DTD and sDTD to model the dependence on the temporal domain. Nevertheless, LSTMs and GRUs are still not powerful enough for action recogntion in practice. One possible reason is that they have only feedforward connections, which is different from biological neural network that is typically composed of both feedforward and feedback connections. Therefore, this paper proposes a biologically-inspired deep network, called shuttleNet, by introducing loop connections in the network. We will show the shuttleNet outperforms LSTMs and GRUs remarkably for action recognition.


\section{Background}\label{sec:background}

\begin{figure}
	\centerline{\includegraphics[width=0.4\textwidth]{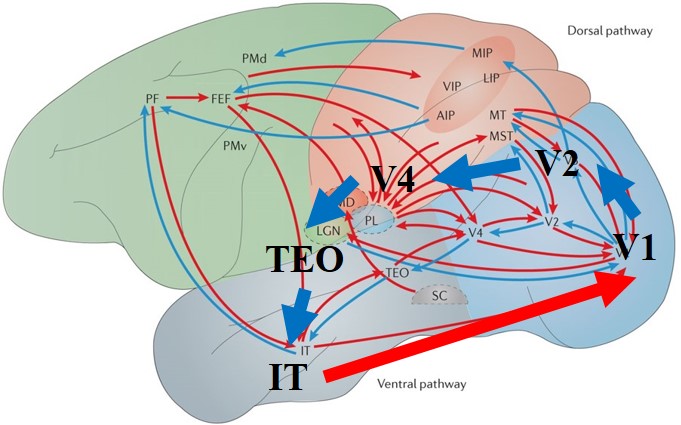}}
	\caption{Illustration of the visual cortical pathways \cite{kandel2000principles}. Feedforward connections are represented by blue arrows and feedback connections are represented by red arrows. A pathway is enlarged for convenience. The feedforward and feedback connections together generate a loop connection.}
	\label{figure:feedback}
\end{figure}

Hierarchical processing is the key to understand vision system. It consists of hierarchically organized distinct anatomical areas functionally specialized for processing different aspects of a visual object \cite{felleman1991distributed}. These visual areas are interconnected through ascending feedforward connections, descending feedback connections and connections from neural structures at the same hierarchical level \cite{lamme1998feedforward}. The lateral geniculate nucleus (LGN) is a relay center in the thalamus for the visual pathway. There is evidence that only 10\% of inputs to LGN come from the retina and 90\% are feedback modulatory inputs from cortex and the brainstem. As shown in Figure \ref{figure:feedback}, the feedback connections play an important role in visual cortical pathways.

\begin{table}
	\centering
	\begin{threeparttable}
		\caption{Some popular deep networks and their depths.}
		\begin{tabular}{c|c|c|c}
			\hline\hline
			LeNet-5 & VGG-16 & GoogLeNet & ResNet-152 \\
			\hline
			5 & 16 & 22 & 152 \\
			\hline\hline
		\end{tabular}
		\begin{tablenotes}
			\footnotesize{\item Please refer to \cite{lecun1998gradient,simonyan2014very,szegedy2015going,he2015deep}.}
		\end{tablenotes}
		\label{table:network_and_depth}
	\end{threeparttable}
\end{table}

On the contrary, DNN researchers concentrate on designing deeper network with only feedforward connections, as shown in Table \ref{table:network_and_depth}. As the model becoming more and more deeper, it will consume more time and computing resource to train. However, LGN and V1 have only 6 layers but still are powerful enough for understanding vision information. A reasonable way to re-design the networks is to imitate the feedback connections of visual cortical pathways.

A feedback connection is a connection along which the information can go back to the previous layer after a few feedforward steps. Together with the feedforward connections, the feedback connections can always result in circular paths. This means that it is not necessary to explicitly generate the feedback connections and we can introduce feedback connections to neural network by generating loop connections.

\section{shuttleNet}\label{sec:shuttleNet}

In this section, we will first present the overall framework of the proposed shuttleNet. After that, we will describe the details about its key components one by one.

\subsection{The overall framework}

\begin{figure*}
	\centerline{\includegraphics[width=0.8\textwidth]{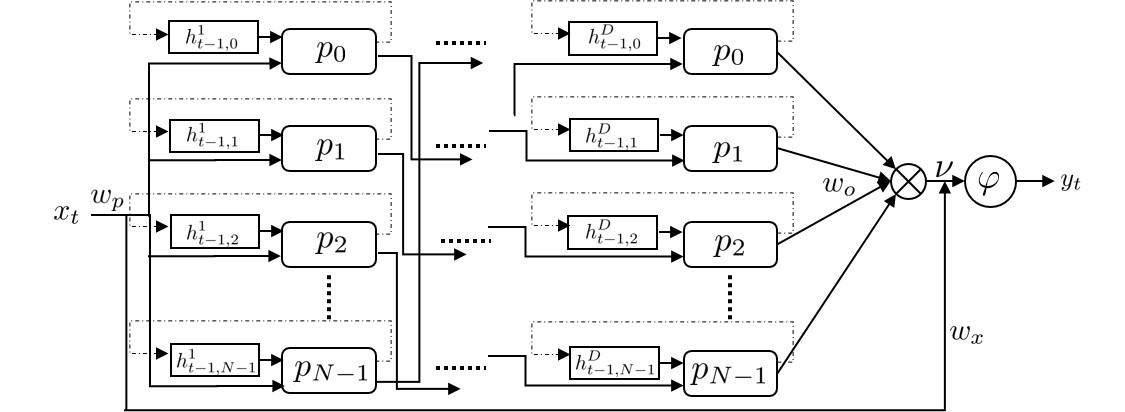}}
	\caption{Diagram of the shuttleNet. The processors are represented by $p_n$ and the cells and states are represented by $h_{t,n}^d$. The information flow pathways are represented by the arrow lines. Every column is called one step. The input $x_t$ is first projected by $w_p$. The projected input is fed into all the $N$ processors and pass through $D$ steps with a stride of $K$, consequently resulting in $N$ pathways. At every step, all processors work simultaneously and generate their outputs. During the steps, all processors are shared while keeping standalone cells and states. Finally, attention mechanism is applied to select the best pathway based on outputs at the last step and input.}
	\label{figure:shuttleNet_new}
\end{figure*}

\begin{figure}
	\centerline{\includegraphics[width=0.45\textwidth]{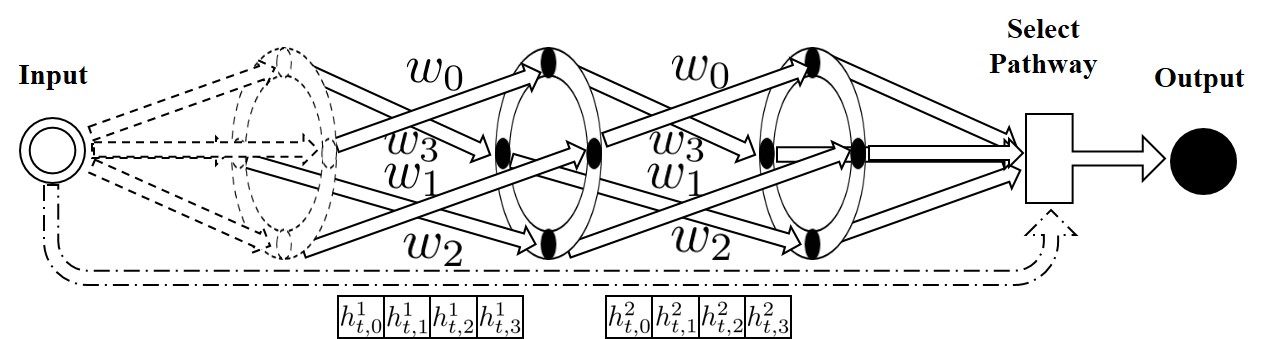}}
	\caption{Illustration of a 4-processor-2-step shuttleNet.}
	\label{figure:shuttleNet}
\end{figure}

Basically, shuttleNet is a biologically-inspired deep neural network, which introduce loop connection and processor-sharing mechanism to the network. As shown in Figure \ref{figure:shuttleNet_new}, the input $x_t$ at time $t$ is projected by a fully-connected layer so that the inputs to processors have the same length as the states. The main module of shuttleNet consists of $N$ processors. The projected input is fed into all the $N$ processors and pass through $D$ steps with a stride of $K$, consequently resulting in $N$ pathways. At every step, all processors work simultaneously and generate their outputs. During these steps, all processors are shared while having standalone cells and states. Finally, attention mechanism is applied to select the best pathway based on outputs at the last step and input.

A noticeable advantage of shuttleNet is that, even though having complex connections, no extra parameters are needed except for the attention module. Considering that the number of parameters does not increase obviously, we can effectively lower the risk of over-fitting.

Our model works like recurrent neural network, and consists of three modules: an input projector, a group of loop-connected processors and an output selector. The entire model (all three components) is trained via backpropagation through time, receiving gradients from every time step, which are then propagated through the unrolled network.

\subsection{Input projector}\label{sec:input}
The input projector is crucial when the inputs and hidden states have different lengths. It makes sure the inputs of processors and hidden states have the same number of dimensions. However, when the input feature length is equal to the length of hidden state, the input projector is not necessary. Typically, the input at time $t$ is a fixed length feature, e.g. the output of CNNs or a bag-of-the-words (BoW) representation. In this work, the inputs are outputs of the last convolution or pooling layer of CNNs. We use a simple fully-connected layer with batch normalization \cite{ioffe2015batch} to project the input. More precisely, let the input be $x_t$. The projected representation of this input is computed as follows:
\begin{align}
x'_t&=w_p x_t \\
x^o_t&=max(\frac{x'_t-E[x'_t]}{\sqrt{Var[x'_t]}}+b, 0)
\end{align}
where $w_p$ and $b$ are the learnable parameters, $E[x]=\frac{1}{N}\sum_{i=1}^N x_i$,  $Var[x]=\frac{1}{N}\sum_{i=1}^{N}(x_i-E[x])^2$, $N$ is batch size and $max(x,0)$ is the ReLU \cite{nair2010rectified} activation function.

\subsection{Loop-connected processors}\label{sec:processor}

As discussed in section \ref{sec:background}, the visual pathways have more feedback connections than feedforward connections. Our first intuition is to design a network with feedforward and feedback connections like the visual pathways. However, to make the model computable and stable, we choose to implement the connections as a simplified version, loop connections. We will then describe it with mathematical linguistic forms.

For most known network structure, layers are stacked one by one and the last layer produces the final output. The layers of a network can be considered as some nonlinear functions. For input $x$, a four-layer network generate the output as follows:
\begin{equation}
y=f_4(f_3(f_2(f_1(x))))
\end{equation}
where $f_1$, $f_2$, $f_3$ and $f_4$ are the four layers and $y$ is the output. In this setting, the network projects input from its original space, where it is hard to classify the samples, to another space, where it is easier to classify the samples. Basically, a network can be decomposed into several nonlinear functions. There are a lot of redundancy in network and it is possible that we share some functions while achieving comparable or better performance as follows:
\begin{equation}
y=f_3(f_1(f_2(f_1(x))))
\end{equation}
There is an obvious advantage in this kind of layer-sharing mechanism that we can decrease number of parameters while keeping the network depth consequently reducing over-fitting.

In order to tap into the potentials of layer sharing, we further develop this idea. A group of layers are organized into a ring. The input $x$ is first fed into all these layers. Then the outputs of the layers are clockwise fed into the next layer with a stride of $K$.

Formally, for $N$ layers $\{f_0, f_1, ..., f_{N-1}\}$ and a given input $x$, we compute $N$ outputs as follows:
\begin{equation}
y_j=f_{(j+D*K)\%N}(...(f_{(j+K)\%N}(f_j(x))))
\end{equation}
where $j\in \{0,1,\dots,N-1\}$, $\%$ is the modulus operator, $D$ is the user defined maximum computing steps and $y_j$ is the output of the $j^{th}$ pathway. Therefore, there are totally $N$ outputs. Because that each layer can be reused multiple times in each pathway consequently resulting in multiple information flow loops and layers are organized into a circle, we call it loop connection.

For a specific application like action recognition, we will implement the layers with recurrent neural networks (RNNs), e.g. LSTMs or GRUs. For the convenience of expression, we will also use ''processor`` to denote the RNN layer.

For a RNN layer, the cell and state $h$ is used to remember history information. The $n^{th}$ RNN processor $p_n$ at $d^{th}$ step and time $t$ in the loop connection works as follows:
\begin{align}
o_{t,n}^d,h_{t,n}&=p_{n}(o_{t,n-K}^{d-1},h_{t,n}) \label{equ:state_1} \\
o_{t,n}^{d+1},h_{t,n}&=p_{n}(o_{t,n-K}^{d},h_{t,n}) \label{equ:state_2}
\end{align}
where $o_{t,n}^d$ is the output of the $n^{th}$ processor at $d^{th}$ step and time $t$ and $h_{t,n}$ is the cell and state of the $n^{th}$ processor. The equation \ref{equ:state_1} means that $h_{t,n}$ is used by $p_n$ at the $d^{th}$ step while equation \ref{equ:state_2} means that $h_{t,n}$ is also used by $p_n$ at the $(d+1)^{th}$ step. However in the loop connection, the two equations are happening in two different information pathways. The chaotic dependencies will end up with unstable training process.

Our solution of the chaotic dependencies is to keep an individual hidden state for each processor at every step while sharing the processor itself between steps. Formally, the processor $p_n$ at $d^{th}$ step and time $t$ in the modified loop connection works as follows:
\begin{align}
o_{t,n}^d,h_{t,n}^d&=p_{n}(o_{t,n-K}^{d-1},h_{t-1,n}^d) \label{equ:state_3} \\
o_{t,n}^{d+1},h_{t,n}^{d+1}&=p_{n}(o_{t,n-K}^{d},h_{t-1,n}^{d+1}) \label{equ:state_4}
\end{align}
where $h_{t,n}^d$ is cell and state of $p_n$ at $d^{th}$ step and time $t$. In these equations, each group of cell and state is only used in one pathway and transmits along time.

\subsection{Output selector}\label{sec:output}
According to section \ref{sec:input} and \ref{sec:processor}, given input $x_t$ at time $t$, after going through the input projector and $N$ pathways, there are $N$ outputs $\{o_{t,0}^D,o_{t,1}^D,\cdots,o_{t,N-1}^D\}$. We need to choose the output which will produce the best prediction. In this paper, attention mechanism \cite{bahdanau2014neural} is employed to help select the proper output. Specifically, the output is computed using the following equations:
\begin{align}
e_{t,n}&=\nu^T tanh(w_x x^o_t+w_o o_{t,n}^D) \\
\alpha_{t,n}&=\frac{exp(e_{t,n})}{\sum_{l=0}^{N-1}exp(e_{t,l})} \\
y_t&=\sum_{l=0}^{N-1}\alpha_{t,l}o_{t,l}^D
\end{align}
where vector $\nu$ and the weight matrixes $w_x$ and $w_o$ are the learnable parameters and the $y_t$ is the output of shuttleNet. The vector $e_t$ assigns a weight for each output $o_{t,n}^D$, which means how much attention should be put on $o_{t,n}^D$. These attention weights are normalized by softmax to create the attention mask $\alpha_t$ over the outputs.

The 3D structure of a 4-processor-2-step shuttleNet with $K=1$ is shown in Figure \ref{figure:shuttleNet}. The input is replicated so that each processor has its own input.  And processors are formed as a circle and work at every step simultaneously. The processors have standalone cell and state at every step. The information flow is propagated in a wheel way. Finally, attention mechanism is utilized to select the best pathway.

It should be noted that when $N$ and $D$ are 1, shuttleNet is completely equivalent to a RNN layer. When $N$ is larger than 1 and $D$ is 1, shuttleNet can be seen as a bank of RNN layers. When $N$ is 1 and $D$ is larger than 1, shuttleNet is a weight-shared stacked RNN.

\section{Experiments}\label{sec:experiments}

In this section, we will first introduce the detail of datasets and their corresponding evaluation schemes. Then, we describe the implementation details of our model. To find out the effect of each parameter, we explore experiments with multiple parameter settings and prove the effectiveness of shuttleNet. We finally report the experimental results and compare shuttleNet with the state-of-the-art methods to demonstrate its superior performance.

\subsection{Datasets}

To verify the effectiveness of shuttleNet, we conduct experiments on two benchmark datasets: HMDB51 \cite{kuehne2011hmdb} and UCF101 \cite{soomro2012ucf101}.

The HMDB51 dataset is a large collection of realistic videos from various sources, including movies and web videos. It is composed of 6,766 video clips from 51 action categories, with each category containing at least 100 clips. The action categories include simple facial actions, general body movements and human interactions. Our experiments follow the original evaluation scheme, and average accuracy over the three train-test splits is reported.

The UCF101 dataset is one of the most popular action recognition benchmarks. It contains 13,320 video clips  (27 hours in total) from 101 action classes and there are at least 100 video clips for each class. The 101 classes are divided into five groups: Body-Motion, Human-Human Interactions, Human-Object Interactions, Playing Musical Instruments and Sports. Following \cite{jiang2014thumos}, we conduct evaluations using 3 train/test splits, which is currently the most popular setting in using this dataset. Results are measured by classification accuracy on each split and we report the mean accuracy over the three splits.

Compared with very large datasets used for image classification, the datasets for action recognition is relatively smaller. Therefore, we pre-train our model on the ImageNet dataset \cite{deng2009imagenet}. Unlike pure CNN models \cite{simonyan2014two,wang2016temporal}, which are fully pre-trained on ImageNet, our network has new layers which can not be trained on image datasets. In order to apply shuttleNet to smaller dataset like HMDB51, we transfer the learnt model from UCF101 to HMDB51.

\subsection{Implementation details}

\begin{figure}
	\centerline{\includegraphics[width=0.35\textwidth]{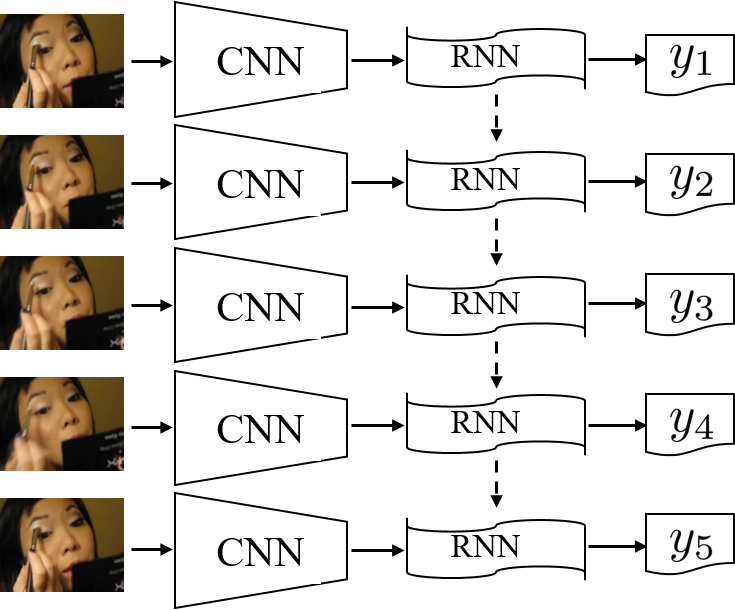}}
	\caption{General action recognition network structure. The video frames are fed into CNNs to learn representations. The CNN representations are then fed into RNNs to learn temporal features. Finally, the output of RNNs is used to predict action label.}
	\label{figure:framework}
\end{figure}

As shown in Figure \ref{figure:framework}, we use a CNN-RNN network structure as in \cite{donahue2015long,shi2017sequential} and use two-stream framework \cite{simonyan2014two} to get our final prediction. In order to conduct as many exploration experiments as we can, we use a relative small network, GoogLeNet \cite{szegedy2015going}, as our CNN implementation to test these parameter combinations. After choosing the best hyper-parameters, we switch to Inception-ResNet-v2 \cite{szegedy2016inception} to get the final performance. To highlight effectiveness of shuttleNet and ensure fairness, the RNN part is tested with several options: 2-layer LSTM \cite{hochreiter1997long}, 2-layer GRU \cite{cho2014learning}, 3-layer GRU and 1-layer shuttleNet. The shuttleNet may have \numrange{1}{3} processors and \numrange{1}{3} steps.

We implement shuttleNet with TensorFlow \cite{abadi2016tensorflow} and use GRU as the processor because of less parameter number and good performance compared with other RNNs.

To extract optical flow, we choose the TVL1 optical flow algorithm \cite{zach2007duality} and use the OpenCV GPU implementation. We discretize the optical flow fields into interval of $[0, 255]$ by a linear transformation and save them as images. Unlike the two-stream ConvNets in \cite{simonyan2014two}, whose temporal stream input is volumes of stacking optical flow fields ($224\times 224\times 2F$, where $F$ is the number of stacking flows and is set to $10$), our temporal stream input is single optical flow. An optical flow field is computed from two consecutive frames and composed of vertical and horizontal flows. To make use of the optical flow, the flow-$x$, flow-$y$ and their quadratic mean is used to form a three channel image.

After initializing with the pre-trained ImageNet model, the network is trained with mini-batch stochastic gradient descent with momentum (set to 0.9). We read 16 frames/optical flows with a stride of 5 from each video as one sample for the RNN. We resize all input images to $256\times 256$, and then use the fixed-crop strategy \cite{wang2015towards} to crop a $224\times 224$ region from images or their horizontal flip. Because the 16 consecutive samples are needed in the RNN, we also force images from the same video to crop the same region. In the test phase, we sample 4 corners and the center from each image and its horizontal flip, and 5 samples are extracted from each video. The batch size is 16 (videos) and the learning rate starts from 0.01. For spatial stream, the learning rate is divided by 10 at iteration $10K$, $20K$ and $30K$, and training is stopped at $40K$ iteration. For temporal stream, the learning rate is divided by 10 at $20K$, $30K$ and $40K$, and training is stopped at $45K$ iteration.

Unlike GoogLeNet, Inception-ResNet-v2's default input size is $299\times 299$. However, such a big input size will result in massive GPU memory requirements. Therefore, we random initialize the auxiliary tower and the classification layer, and then force the Inception-ResNet-v2 to train with $224\times 224$ images. And we use RMSProp to optimize the model. Another issue of training Inception-ResNet-v2 with RNNs is that the batch normalization is highly relied on the mean and variance of each batch. When training Inception-ResNet-v2 with RNNs, each GPU (with 12G memory e.g. one core of Tesla K80) can only hold 32 images, which are two videos. In order to get reasonable mean and variance, we train the model with 8 GPUs, which is capable of 16 videos. Then we select 2 images from each video for each GPU and get 32 images. The hand-selected batches, which have 32 images, are respectively fed into 8 GPUs. After getting features from the last layer of CNNs (or the input projector), we rearrange the images to their original order to train the RNNs (or shuttleNet). Even though it is complicated to train RNNs (or shuttleNet) with Inception-ResNet-v2, the test stage is very easy and we are able to test with a batch size of 160 images without any modification.

In the remainder of the paper, we use spatial stream and temporal stream to indicate the streams in two-stream framework and use ``two streams'' to indicate the late fusion of two models.

\subsection{Exploration experiments}\label{subsubsection:exporation}

In this section, we will first test shuttleNet with different setting by adjusting the number of processors and the number of steps. Then, we compare our model with several existing RNN models in section \ref{subsubsection:comparison_baseline} to prove the effectiveness of shuttleNet. We find that our shuttleNet can achieve much better performance with similar number of parameters. We also select base network by train shuttleNet with GoogLeNet and Inception-ResNet-v2 and compare their accuracy.

\begin{table}
	\centering
	\caption{Accuracy and parameter number of shuttleNet with different settings on UCF101 split 1.}
	\begin{tabular}{c|c|c|c}
		\hline\hline
		\# processors & \# steps & Spatial stream & Params \\
		\hline
		\multirow{3}{*}{2} & 1 & 82.37\% & 10.49M \\
		& 2 & 83.03\% & 10.49M \\
		& 3 & 80.89\% & 10.49M \\
		\hline
		1 & \multirow{3}{*}{2} & 82.9\% & 6.29M \\
		2 & & 83.03\% & 10.49M \\
		3 & & 83.19\% & 14.68M \\
		\hline
		3 & 3 & 82.13\% & 14.68M \\
		\hline\hline
	\end{tabular}
	\label{table:accuracy_different_param}
\end{table}

\begin{table*}
	\centering
	\caption{Accuracy of shuttleNet on two datasets.}
	\begin{tabular}{c|c|c|c|c|c}
		\hline\hline
		Dataset & split & Spatial & Temporal & Two streams & Two streams + MIFS \\
		\hline
		\multirow{4}{*}{UCF101} & 1 & 87.9\% & 86.3\% & 94.1\% & 95.1\% \\
		& 2 & 87.1\% & 88.6\% & 94.6\% & 95.8\% \\
		& 3 & 86.8\% & 87.4\% & 94.3\% & 95.2\% \\
		\cline{2-6}
		& Mean & 87.3\% & 87.4\% & 94.4\% & 95.4\% \\
		\hline
		\multirow{4}{*}{HMDB51} & 1 & 55.6\% & 61.9\% & 69.0\% & 74.2\% \\
		& 2 & 53.0\% & 58.9\% & 66.0\% & 71.2\% \\
		& 3 & 53.9\% & 59.5\% & 64.8\% & 69.8\% \\
		\cline{2-6}
		& Mean & 54.2\% & 60.1\% & 66.6\% & 71.7\% \\
		\hline\hline
	\end{tabular}
	\label{table:acc_all_datasets}
\end{table*}

\begin{table}
	\centering
	\begin{threeparttable}
		\caption{Comparison of shuttleNet to several popular recurrent neural networks on UCF101 split 1. The second column is the number of the processors. The shuttleNets use 2 steps.}
		\begin{tabular}{c|c|c|c|c}
			\hline\hline
			Model & \# & Spatial & Temporal & Two Streams \\
			\hline
			GoogLeNet\textdagger & - & 77.74\% & 79.75\% & 85.91\% \\
			LSTM & 2 & 78.8\% & 80.15\% & 88.58\% \\
			GRU & 2 & 80.49\% & 82.77\% & 90.27\% \\
			GRU & 3 & 82.42\% & 81.21\% & 90.64\% \\
			\hline
			shuttleNet & 1 & 82.9\% & 82.92\% & 91.65\% \\
			shuttleNet\textdaggerdbl & 2 & 83.03\% & 82.18\% & 91.94\% \\
			\hline\hline
		\end{tabular}
		\begin{tablenotes}
			\footnotesize{\item[\textdagger] There is no RNN in GoogLeNet \cite{szegedy2015going}.
				\item[\textdaggerdbl] We use 1 group of optical flow fields as one sample for GoogLeNet instead of 10 as in \cite{simonyan2014two,wang2016temporal}, which result in relatively worse performance than others'. But we can still get state-of-the-art result based on this implementation.}
		\end{tablenotes}
		\label{table:comparison_lstm}
	\end{threeparttable}
\end{table}

\begin{table}
	\centering
	\caption{Number of parameters in 2-layer RNN models with state size of 1024 and 1-layer shuttleNet without the input projector.}
	\begin{tabular}{c|c|c|c}
		\hline\hline
		Model & LSTM & GRU & shuttleNet \\
		\hline
		Params & 16.78M & 8.39M & 10.49M \\
		\hline\hline
	\end{tabular}
	\label{table:number_param}
\end{table}

\begin{table}
	\centering
	\caption{The mean accuracy of shuttleNet with GoogLeNet and Inception-ResNet-v2 on the three splits of UCF101.}
	\begin{tabular}{c|c|c|c}
		\hline\hline
		& Spatial & Temporal & Merge \\
		\hline
		GoogLeNet & 81.4\% & 84.9\% & 92.3\% \\
		\hline
		\tabincell{c}{Inception-\\ResNet-v2} & 87.3\% & 87.4\% & 94.4\% \\
		\hline\hline
	\end{tabular}
	\label{table:sNet_net}
\end{table}

\subsubsection{Parameter study}

%

There are three hyper parameters in shuttleNet: the stride $K$, the number of processors $N$ and the number of steps $D$. We will fix $K$ as 1 in all of our experiments and explore the effect of other parameter by running experiments under multiple settings.

The performance of shuttleNet under different settings are shown in table \ref{table:accuracy_different_param}. We find that as the number of steps growing, the performance of spatial stream first increases then decreases. As the number of processors growing, the spatial stream accuracy keeps increasing and parameter number grows quickly. Actually, the conclusions are easy to understand. When increasing the number of steps, each processor will have to take care of more situations. The more steps used, the harder for processors to learn. However, with 2 steps, it works like regularization approach preventing overfitting hence improving the performance. When increasing the number of processors, all processors work like model ensemble. The more processors used, the better the ensemble will be. However, more processors will result in more parameters hence increasing complexity.

Based on the above discussion, we will use 2 steps in the following sections. Although our shuttleNet can be easily extended as a multi-layer model, we use only one layer shuttleNet with 2 processors to compare with other models, for example, 2-layer LSTM or GRU, to emphasize the good performance and ensure fairness.

\subsubsection{Comparison with baselines}\label{subsubsection:comparison_baseline}

As shown in Table \ref{table:comparison_lstm}, we compare the shuttleNet with several popular recurrent neural network models. The shuttleNet achieves a good performance of 91.94\% on UCF101 split 1 while none of other models close to 91\%. On the spatial stream, shuttleNet outperforms other RNN models remarkably. Even the three layer GRU network can not beat our 2-processor shuttleNet.

Unlike most of existing methods, who use a stack of 10 optical flow fields as one sample, all of our experiments use 1 group of optical flow fields as one sample to reduce complexity. Therefore we get relatively worse performance on the temporal stream and two-stream model. But we can still achieve the state-of-the-art performance after using shuttleNet. We believe we can boost the performance by using stack of 10 optical flow fields as input regardless of the complexity.

To prove the effectiveness of output selector, we conduct an experiment with a 2-processor-1-step shuttleNet. The accuracy of spatial stream is 82.37\% and temporal stream accuracy is 82.76\%. The two-stream shuttleNet achieves an accuracy of 91.7\%. This means that output selector is very effective and can improve the performance significantly.

It's amazing that our 1-processor-2-step shuttleNet also achieve an accuracy of 91.65\%, which is still much better than all other RNN models. The great performance of 1-processor shuttleNet proves the effectiveness of our processor-sharing mechanism.

The processor-sharing mechanism also makes sure that our model won't have too many parameters. As shown in Table \ref{table:number_param}, our model has slightly more parameters than GRU while having much less parameters than LSTM. The addition parameters are used to compute the attention value for each path. This means that no matter how many processors we have, there won't be any more parameters needed.

The overall potential of loop connection is fully demonstrated by the 2-processor-2-step shuttleNet. It outperforms all of these baselines and achieves a high accuracy of 91.94\%. The good performance proves the advantage of shuttleNet.

Parallel computing is another potential advantage of our shuttleNet. Because that each processor in shuttleNet at every step works independently, it's easy to accelerate the computing by parallel computing all processors.

\subsubsection{Selection of base network}

The shuttleNet gets inputs from previous CNNs. It is important to select a better network. We test shuttleNet with GoogLeNet and Inception-ResNet-v2 and the results are shown in Table \ref{table:sNet_net}. The Inception-ResNet-v2 gets a 2\% higher accuracy than GoogLeNet. The shuttleNet achieves state-of-the-art performance even using a relatively simpler network like GoogLeNet. And it achieves much better performance by using Inception-ResNet-v2. In the rest of the paper, we will use Inception-ResNet-v2 as our base network and report the accuracies.

\subsection{Evaluation}

In this section, we will first evaluate shuttleNet on more datasets and report their accuracy. Then we compare shuttleNet with several state-of-the-art methods in section \ref{sec:comp_state_of_the_art}.

\subsubsection{Evaluation of shuttleNet}\label{subsubsection:evalution}

From section \ref{subsubsection:comparison_baseline}, we prove the effectiveness of shuttleNet on UCF101 split 1. We now report the performance of shuttleNet on more splits and dataset with Inception-ResNet-v2 in Table \ref{table:acc_all_datasets}.

For the HMDB51 dataset, shuttleNet performs good on the first split while relatively worse on the other two splits. And the overall performance is not as good as in UCF101. We believe this is because of the small training set size (3570 videos). But shuttleNet is still outstanding on HMDB51 when comparing to other methods. On UCF101 dataset, the shuttleNet performs good on all three splits.

In most cases, the temporal stream achieves better performance than the spatial stream. This proves that the motion information plays an important role in action recognition.

Compared to pure CNN models, our method adds new layers to the network, which can not be pre-trained on ImageNet. Training the new layers on small dataset, e.g. HMDB51, is easy to be over-fitting. In order to apply shuttleNet to small dataset, we merge shuttleNet with MIFS \cite{lan2015beyond}, which is a hand-crafted feature, using late fusion. The improvement proves the complementation between deep features and traditional features.

\subsubsection{Comparison with the state-of-the-art methods}\label{sec:comp_state_of_the_art}

\begin{table}
	\centering
	\caption{\textnormal{Comparison of shuttleNet to the state-of-the-art methods.}}
	\begin{tabular}{c|c|c}
		\hline\hline
		Model & HMDB51 & UCF101 \\
		\hline
		STIP+BoF \cite{kuehne2011hmdb} & $23.0\%$ & $43.9\%$ \\
		DT+MVSV \cite{cai2014multi} & $55.9\%$ & $83.5\%$ \\
		iDT+FV \cite{wang2013action} & $57.2\%$ & $85.9\%$ \\
		iDT+HSV \cite{peng2016bag} & $61.1\%$ & $88.0\%$ \\
		MIFS \cite{lan2015beyond} & $65.1\%$ & $89.1\%$ \\
		Two-stream ConvNets \cite{simonyan2014two} & $59.4\%$ & $88.0\%$ \\
		F$_{ST}$CN \cite{sun2015human} & $59.1\%$ & $88.1\%$ \\
		TDD+FV \cite{wang2015action} & $63.2\%$ & $90.3\%$ \\
		TDD+iDT+FV \cite{wang2015action} & $65.9\%$ & $91.5\%$ \\
		MoFAP \cite{wang2015mofap} & 61.7\% & 88.3\% \\
		LTC \cite{varol2016long} & 64.8\% & 91.7\% \\
		sDTD \cite{shi2017sequential} & 65.2\% & 92.2\% \\
		Conv Fusion \cite{feichtenhofer2016convolutional} & 65.4\% & 92.5\% \\
		Three-stream TSN \cite{wang2016temporal} & 69.4\% & 94.2\% \\
		\hline
		Ours & $\textbf{71.7\%}$ & $\textbf{95.4\%}$ \\
		\hline\hline
	\end{tabular}
	\label{figure:compare_state_of_the_art}
\end{table}

Finally, we compare against the state-of-the-art over all three splits of HMDB51 and UCF101. The results are summarized in Table \ref{figure:compare_state_of_the_art}, where we compare our method with both traditional approaches such as iDT and deep learning methods such as TSN \cite{wang2016temporal}.

Compared to the two-stream ConvNets \cite{simonyan2014two}, which is the most famous baseline method, we get around  12.3\% and 7.4\% improvements on HMDB51 and UCF101 datasets, respectively. These results are much better than traditional iDT \cite{wang2013action} and MIFS \cite{lan2015beyond}. Compared to TSN \cite{wang2016temporal}, the shuttleNet achieves better performance on both HMDB51 and UCF101.

Even though we use a very simple framework and only replace RNNs with our shuttleNet, our model still outperforms these methods on UCF101 and HMDB51. The superior performance of our method demonstrates the effectiveness of shuttleNet. It's possible that we can still improve the performance by using 10 groups of optical flow fields as input or pre-train the new layers of shuttleNet on large dataset like YouTube-8M \cite{abu2016youtube}.

\section{Conclusion}\label{sec:conclusion}
In this paper, we present a biologically-inspired deep network, shuttleNet, for action recognition. As demonstrated on two benchmark datasets, shuttleNet can outperform LSTMs and GRUs remarkably while having roughly equal number of parameters. This is largely ascribed to the loop connections as well as processor-sharing mechanism. In the future, there are still a lot of research directions that will be addressed, for example, to check the performance of the stacked shuttleNet and whether CNNs will benefit from loop connection.

{\small
\bibliographystyle{ieee}
\bibliography{egbib}
}

\end{document}